\pdfoutput=1

\documentclass[11pt]{article}

\usepackage{acl}

\usepackage{times}
\usepackage{latexsym}
\usepackage{framed}

\usepackage[T1]{fontenc}

\usepackage[utf8]{inputenc}

\usepackage{microtype}

\usepackage{inconsolata}

\usepackage{graphicx}
\usepackage{tcolorbox}
\usepackage{multirow}
\usepackage{graphicx}
\usepackage{booktabs}
\usepackage{array}
\usepackage{makecell}
\usepackage[utf8]{inputenc} 
\usepackage[T1]{fontenc}    
\usepackage{booktabs}
\usepackage{amsmath} 
\usepackage{siunitx} 
\usepackage{url}            
\usepackage{booktabs}       
\usepackage{amsfonts}       
\usepackage{nicefrac}       
\usepackage{microtype}      
\usepackage{graphicx}
\usepackage{booktabs}
\usepackage{graphicx}
\usepackage{amsmath, amsthm, amssymb}
\usepackage{algorithm}
\usepackage{algorithmic}
\usepackage{cleveref}
\usepackage{booktabs}
\usepackage{multirow}
\usepackage{colortbl}
\usepackage{collcell}
\usepackage{arydshln}
\usepackage{tabularx}
\usepackage{lipsum}
\usepackage{wrapfig}
\usepackage{pifont}
\usepackage{capt-of}
\usepackage{bm}
\usepackage{multirow}
\usepackage{graphicx}
\usepackage{lscape}
\usepackage{amsmath}
\usepackage{amssymb}
\usepackage{amsthm}
\usepackage{graphicx}
\usepackage{hyperref}
\usepackage{listings}
\usepackage{booktabs}
\usepackage{makecell}
\usepackage{lipsum} %
\usepackage[hang,flushmargin]{footmisc} 

\definecolor{darkgreen}{rgb}{0.0, 0.5, 0.0} 
\definecolor{darkred}{rgb}{0.5, 0.0, 0.0}   

\newcommand{\OUR}{PaCoST}

\newcommand{\cmark}{\textcolor{darkgreen}{\ding{51}}}%
\newcommand{\xmark}{\textcolor{red}{\ding{55}}}%

\newcommand\blfootnote[1]{\begingroup\renewcommand\thefootnote{}\footnote{#1}\addtocounter{footnote}{-1}\endgroup}

\definecolor{linkcolor}{HTML}{ED1C24}

\definecolor{color1}{HTML}{ffdebf}
\definecolor{color2}{HTML}{ffefe0}
\definecolor{color3}{HTML}{E6ECE3}
\definecolor{color4}{HTML}{feeafa}
\definecolor{color5}{HTML}{dee2ff}
\definecolor{color6}{HTML}{ffc0c3}
\definecolor{color7}{HTML}{00ffff}

\title{PaCoST: Paired Confidence Significance Testing for Benchmark Contamination Detection in Large Language Models}

\author{Huixuan Zhang$^{1,*}$\quad Yun Lin$^{1,2,*}$\quad Xiaojun Wan$^{1}$  \\
\textsuperscript{\rm 1}   Wangxuan Institute of Computer Technology, Peking University\\
\textsuperscript{\rm 2}  School of Foreign Languages, Peking University \\
\texttt{\{zhanghuixuan,linyun\}@stu.pku.edu.cn,wanxiaojun@pku.edu.cn}
}

\begin{document}
\maketitle
\blfootnote{$^{*}$ Both authors contributed equally to this research.}

\begin{abstract}

Large language models (LLMs) are known to be trained on vast amounts of data, which may unintentionally or intentionally include data from commonly used benchmarks. This inclusion can lead to cheatingly high scores on model leaderboards, yet result in disappointing performance in real-world applications. To address this benchmark contamination problem, we first propose a set of requirements that practical contamination detection methods should follow. Following these proposed requirements, we introduce \textbf{PaCoST}, a \underline{Pa}ired \underline{Co}nfidence \underline{S}ignificance \underline{T}esting to effectively detect benchmark contamination in LLMs. Our method constructs a counterpart for each piece of data with the same distribution, and performs statistical analysis of the corresponding confidence to test whether the model is significantly more confident under the original benchmark. We validate the effectiveness of PaCoST and apply it on popular open-source models and benchmarks. We find that almost all models and benchmarks we tested are suspected contaminated more or less. We finally call for new LLM evaluation methods. \footnote{\space Our code will be released at \url{https://github.com/lleozhang/PaCoST}.} 
\end{abstract}

\section{Introduction}

Large Language Models (LLMs) have brought about a paradigm shift in the domain of natural language processing, yielding notable enhancements across various evaluation benchmarks \cite{wang2019superglue} and demonstrating proficiency in professional examinations\cite{openai2023gpt4}. These advancements primarily stem from extensive training on vast and diverse datasets sourced from multiple origins. However, the substantial volume of data has given rise to significant concerns regarding benchmark contamination, where benchmarks for LLM evaluation are inadvertently or deliberately included in model training. This contamination presents considerable obstacles in accurately gauging the capabilities of LLMs. 



 While efforts are being made to address this issue by removing benchmarks from training datasets and conducting contamination studies, these endeavors face numerous limitations \cite{brown2020language, zhang2024careful, wei2022finetuned, chowdhery2022palm}. These limitations include narrow focus on specific benchmarks and reliance on the trustworthiness of vendors. Moreover, the competitive dynamics within the field, coupled with copyright considerations, have resulted in recent model releases lacking accompanying contamination studies \cite{openai2023gpt4}. Hence, there is an urgent necessity for independent methods to audit LLMs for the presence of benchmark datasets, eliminating the dependence on model providers' cooperation.

Simultaneously, there has been a growing interest in heuristic membership inference algorithms designed to reverse-engineer aspects of the training dataset \cite{carlini2021extracting, mattern-etal-2023-membership}, thereby providing insights into potential test set contamination \cite{sainz2023nlp, golchin2023time}. Despite their promise, these heuristic approaches often lack definitive proof of contamination and tend to rely on assumptions that may be too stringent. Moreover, the majority of these methods concentrate less on detecting benchmark contamination. As elaborated in Section \ref{sec:3}, inherent challenges, such as the need for lengthy trained segments and the necessity of establishing thresholds, impede the adaptation of previous methods for detecting benchmark contamination.

In this study, we introduce a novel approach named \textbf{PaCoST} (\underline{Pa}ired \underline{Co}nfidence \underline{S}ignificance \underline{T}esting) designed for the detection of benchmark contamination in open-source LLMs. Our method entails a three-step statistical analysis, capable of identifying benchmarks within the model's training data. Specifically, our approach involves constructing counterparts for each data instance with similar distribution, followed by statistical analysis of corresponding confidence scores to ascertain whether the model exhibits significantly higher confidence when presented with original benchmarks. We operate under the assumption that the model tends to demonstrate greater confidence when responding to questions it has been trained on. To validate our method rigorously, we conduct a series of controlled experiments.

Subsequently, we employ PaCoST across a diverse array of publicly accessible LLMs, scrutinizing various benchmarks to reveal contamination outcomes. Our experimental observations indicate that, across the board, there are suspicions of contamination to varying degrees in both models and benchmarks. Consequently, we advocate for the adoption of a benchmark-free evaluation approach as a means to mitigate this contamination issue.

Our contributions can be summarized as follows:
\begin{itemize}
    \item We propose several properties which a good benchmark contamination detection method should satisfy.
    \item We introduce a simple yet effective method PaCoST to detect benchmark  contamination in LLMs and validate its effectiveness and stability.
    \item We conduct experiments on popular open-source LLMs and benchmarks and find suspected contamination on almost all tested models and benchmarks. 
\end{itemize}

\section{Related Works}

\subsection{Data Contamination Detection}
The issue of data contamination in large language models has been increasingly recognized as a significant concern \citep{sainz2023nlp}. Many LLM providers use string-matching to report contamination, such as GPT-2 \cite{radford2019language}, GPT-3 \cite{gpt3}, PaLM \cite{chowdhery2023palm}, GPT-4 \cite{openai2023gpt4}, and Llama 2 \cite{touvron2023llama2}. However, in most cases, the model's training data is not publicly available, necessitating alternative detection methods.

Several methods have been developed to detect data contamination in LLMs. \citet{nasr2023scalable} and \citet{did-ChatGPT-cheat-on-your-test} explore the regeneration of initial dataset instances. \citet{golchin2023time} introduces guided prompting to replicated trained data. \citet{golchin2023data} develops a Data Contamination Quiz (DCQ) framework. 

Beyond prompt-based methods, there are also methods based on likelihood such as the Min-K\% Prob \cite{shi2024detecting}, \citet{oren2023proving} and \citet{li2023estimating}. Additionally, methodologies like CDD and TED \citep{dong2024generalization} focus on the LLM’s output distribution. But these methods do not pay enough attention to benchmark contamination detection.

Membership Inference Attack (MIA) is closely related to data contamination, aiming to identify whether a given sample is in a model's training data \citep{shokri_membership_2017-1, yeom_privacy_2018}. These attacks pose significant privacy risks and can lead to severe breaches \cite{carlini_extracting_2021, gupta_recovering_2022, cummings_challenges_2023}. MIA is crucial for assessing privacy vulnerabilities and validating privacy-preserving measures in machine learning models \cite{jayaraman_evaluating_2019, jagielski_auditing_2020, nasr2023scalable}. Initially applied to tabular and computer vision data, MIA has recently been extended to language-based tasks \cite{song_auditing_2019, shejwalkar_membership_2021, mahloujifar_membership_2021, mireshghallah_quantifying_2022}.

\begin{table*}[t]
    \footnotesize
    \centering
    \setlength{\tabcolsep}{4mm} 
    \begin{tabular}{l|ccccc}
        \toprule
        Method&  TDA Free & CT Free & TDL Free & SP & T Free \\
        \midrule
        String-match \citep{openai2023gpt4} &\xmark & \cmark & \cmark & \cmark & \cmark \\
 Min-k\% Prob \citep{shi2024detecting}& \cmark & \xmark & \xmark &\cmark & \xmark\\
        Guided-Prompting  \citep{golchin2023time}&\cmark & \xmark & \xmark & \xmark & \cmark \\
        Sharded-Likelihood \citep{oren2023proving} &\cmark & \xmark & \xmark & \cmark & \xmark \\
        CDD \citep{dong2024generalization}&\cmark & \xmark & \xmark & \xmark & \xmark\\
        
        DCQ \citep{golchin2023data}& \cmark & \cmark & \cmark & \xmark & \cmark \\
        \midrule
        \textbf{\OUR} (ours) &\cmark & \cmark & \cmark & \cmark & \cmark\\
        \bottomrule
    \end{tabular}
    \caption{Comparison of existing methods and \OUR. \cmark\space means the method satisfies the corresponding property and \xmark \space refers to methods not satisfying the corresponding property. The name of properties are abbreviated for presentation and their full contents can be found in Section \ref{sec:prop}.}
    \label{tab:method-compare}
\end{table*}

\subsection{Confidence Estimation}
Estimating the confidence of a model in its output is a critical challenge in the research of LLMs.  \citet{kuhn2023semantic} aggregates probabilities of semantically equivalent answers to determine confidence. Other methods include directly querying the model for its confidence \cite{lin2022teaching,tian2023just} and calculating self-consistency scores \cite{wang2022self,lin2023generating}. Some techniques for confidence calibration involve modifying prompts and paraphrasing instructions to fine-tune the probability distribution \citep{zhao2023knowing, jiang2023calibrating}, or using the probability that the model agrees with its own answers, such as in P(True) \citep{kadavath2022language}. Combined approaches further enhance calibration accuracy \cite{xiong2023can, chen2023quantifying}.

\section{Problem Formulation}
\subsection{Benchmark Contamination}
\label{sec:3}
In this study, we focus on detecting benchmark contamination. The problem is formulated as: consider a benchmark $D=\{(x_1, y_1), ..., (x_n, y_n)\}$, where $x_i$ denotes an instruction and $y_i$ represents the ground truth answer. We define benchmark contamination as the model has been trained to maximize $\mathcal{P}(y_i|x_i)$ (or to minimize $-\log \mathcal{P}(y_i|x_i)$).

There are two contamination types that align with this objective. For a given data instance $(x,y)$, the first contamination type performs next-token prediction on both the instruction $x$ and the answer 
$y$, which aims at minimizing:
\begin{align}
\nonumber
-\log \mathcal{P}(x,y) &= -\log \mathcal{P}(y|x)\mathcal{P}(x)\\
\nonumber
&= -(\log \mathcal{P}(y|x) + \log \mathcal{P}(x))
\end{align}
The second contamination type only performs next-token prediction on the answer $y$, which aims at minimizing $-\log \mathcal{P}(y|x)$. The only difference between the two contamination types lies in whether $-\log \mathcal{P}(x)$ is part of the optimizing objective. 

\subsection{Detection Requirements}
\label{sec:prop}
Building upon the formulation outlined earlier and taking into account the features of existing methodologies for detecting data contamination, we propose several key criteria that a robust benchmark contamination detection method should satisfy.

\paragraph{I. Training Data Access Free (TDA Free)}
While String-Match might offer high accuracy in detecting data contamination, it is frequently impractical due to LLM providers' reluctance to disclose training datasets. Even if training datasets were accessible, the sheer volume of data makes pinpointing specific instances nearly impossible. Hence, reliance on access to original training data for contamination detection is neither feasible nor practical. Effective benchmark contamination detection methods must be engineered to operate independently of training data access.


\paragraph{II. Contamination Type Free (CT Free)}
Many conventional contamination detection methods primarily target the first type of contamination, where both the instruction and answer parts are trained. This focus is reasonable for detecting contamination in unlabeled data. However, benchmark contamination can also manifest in the second type, where only the answer part undergoes training, rendering many existing methods unsuitable for addressing this issue. For example, techniques like Min-k\% Prob \citep{shi2024detecting}, which entails computing the minimum k\% probabilities of the entire input, may fail to function accurately if the instruction part remains untrained. Hence, an effective detection method should not be constrained by contamination type. 


\paragraph{III. Training Data Length Free (TDL Free)}
Building on the preceding discussion, since most data contamination detection methods focus on the first type of contamination, they naturally presume relatively lengthy trained parts. However, benchmark contamination can also occur with only a very short answer part $y$ trained (e.g., merely an option or a word). This renders assumptions about training length invalid, making methods reliant on such assumptions ineffectual. Still taking Min-k\% Prob \citep{shi2024detecting} as an example, if we solely compute the minimum k\% probabilities on the response part, it will introduce excessive noise due to the brevity of the response part. Hence, a robust benchmark contamination detection method should not be constrained by training length. This will be further discussed in Appendix \ref{sec:app-comp}.

\paragraph{IV. Stable Performance (SP)}
Certain methods exhibit sensitivity to prompts or settings, necessitating specific prompts for proper functionality.  Guided-Prompting \citep{golchin2023time} mandates knowledge of the dataset's name and split to construct the guided prompt, which may not always be attainable. Moreover, the disparity between general and guided prompts, even without considering dataset metadata, casts doubt on the method's stability. Similarly, DCQ \citep{golchin2023data} mandates the model to choose from five options, and altering the order of options yields disparate results, rendering its detection outcomes meaningless. Therefore, a robust detection method should yield stable results despite reasonable changes in settings.

\paragraph{V. Threshold Free (T Free)}
Some methods necessitate the selection of a threshold for detection, such as Min-k\% Prob \citep{shi2024detecting}. However, datasets and models exhibit varying distributions, rendering a universal threshold impractical. While some threshold-sensitive methods resort to reporting Area Under the Curve (AUC) for quantitative comparison to circumvent this issue, in real-world scenarios, employing a specific threshold for detection is unavoidable. Therefore, we contend that a threshold-based method should provide a fixed threshold and demonstrate its effectiveness across all datasets rather than relying on AUC. A superior detection method should not entail flexible thresholds; all hyperparameters should be predefined.

We examine the most popular data contamination detection methods and compare them in Table \ref{tab:method-compare}. As evident, all methods, except our proposed one, fail to satisfy all properties. This observation underscores the advantages of our method.

\begin{figure*}[htb]
    \centering
    \includegraphics[width=1\linewidth]{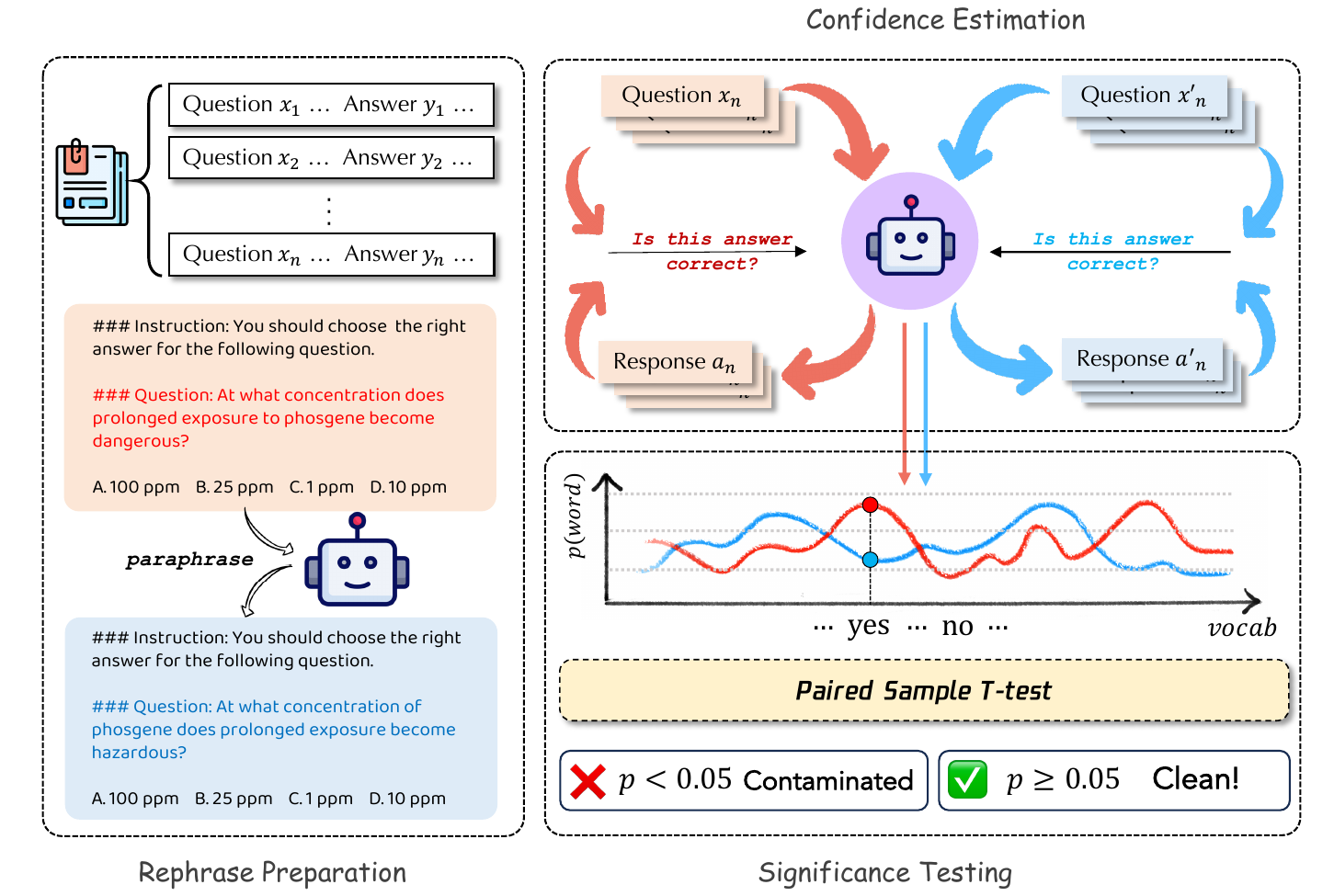} 
    \caption{Overview of our method. $x_i$ represents a question, $y_i$ represents its corresponding ground truth answer, $x_i^{'}$ represents a rephrased question and $a_{i}, a_{i}^{'}$ represent model responses to original and rephrased question correspondingly. }
    \label{fig:method}
\end{figure*}

\section{Method}

We introduce PaCoST, a novel benchmark contamination detection method that emphasizes the distinction between contaminated and clean data without relying on thresholds. Our approach leverages the disparity in model behavior between original and rephrased instances, focusing on confidence rather than traditional performance metrics like accuracy \citep{yang2023rethinking}. By conducting statistical analysis on confidence, we can robustly identify contamination. PaCoST comprises three key steps: rephrasing preparation, confidence estimation, and significance testing. Through this method, we provide a clear and unique approach to detecting benchmark contamination in models.

\subsection{Rephrase Preparation}
Our key idea involves comparing confidence between original and rephrased instances. We opt for rephrasing for several reasons. First, to ensure a fair comparison, the trained and untrained data should share similar distributions and levels of difficulty. Otherwise, comparing confidence would be meaningless. Creating questions with the same distribution and difficulty but different meanings is challenging. Second, rephrasing is a fundamental capability of most common LLMs, making it straightforward to implement.

Given an instance $(x, y)$,  we use a model $M_{p}$ to rephrase $x$ into $x' = M_{p}(x)$ while $y$ remain unchanged. 
We select Llama2-Chat-7B \citep{touvron2023llama2} as the rephrase model for all the tested models (The rephrase prompt is provided in Appendix \ref{sec:paraphrase prompt}). To validate the quality of the rephrasing, we employ both BERT-Score \citep{bert-score} and human annotation. Additionally, we compare the performance of different models for rephrasing and demonstrate that using various paraphrasing models does not impact performance, provided they are sufficiently powerful. Further details can be found in Appendix \ref{sec:app-2}. 

\subsection{Confidence Estimation}

There are various ways to estimate a model's confidence in its answers, as previously discussed. In this study, we select the method P(True) \citep{kadavath2022language} for confidence estimation.

We briefly introduce this method. Consider an instance $(x,y)$, where $x$ is an instruction and $y$ is the ground truth answer. For an LLM $M$ and its corresponding output $M(x)$, P(True) queries the model $M$ whether $M(x)$ is a correct answer to $x$. Denote the output probability distribution of querying as $\mathcal{P}(\cdot |x, M(x), M)$, the confidence can be then denoted as $\mathcal{P}(True|x, M(x), M)$ where True represents model $M$ supporting $M(x)$. According to our setting and prompt, we are actually calculating $\mathcal{P}(Yes|x, M(x), M)$.

We opt for using P(True) for confidence estimation for several reasons. First, using probability distribution of the original output ($\mathcal{P}(M(x)|x, M)$ to estimate confidence often leads to overconfidence issues, resulting in unnaturally high confidence scores \citep{xiong2023can}. This problem also partly explains why methods like Min-k\% Prob are ineffective on relatively short training segments. We will further explore this observation in Appendix \ref{sec:app-comp}.

Second, Verbalized confidence estimation methods, which involve directly querying the model to provide a confidence score, often yield discrete confidence values. This makes them unsuitable for our purposes. Other confidence estimation methods are generally either inappropriate or overly complex. Therefore, we ultimately choose P(True) for its simplicity and effectiveness. Details of our prompt and an example can be found in Appendix \ref{sec:paraphrase prompt}.

\subsection{Significance Testing}

Consider a benchmark 
$D$ $=$ $\{(x_1,y_1), ..., $ $(x_n,y_n)\}$
and its rephrased benchmark 
$
D^{'} = \{(x_1^{'} ,y_1),...,(x_n^{'} ,y_n)\}
$
we have calculated the paired confidence set $\{(c_1, c_1^{'}),...,(c_n, c_n^{'})\}$, where 
\[
c_i = \mathcal{P}(Yes|x_i, M(x_i), M)
\] and 

\[
c_i^{'} = \mathcal{P}(Yes|x_i^{'}, M(x_i^{'}), M)
\] 

We use Paired Samples T-test to perform statistical analysis. Denote $d_{i} = c_{i} - c_{i}^{'}$, assuming $d_i \sim \mathcal{N}(\mu, \sigma^{2})$, we would like to test whether $\mu > 0$. Then the null hypothesis $H_0$ and the alternative hypothesis can be denoted as 
\[
    H_0: \mu \leq 0 \longleftrightarrow H_1: \mu > 0
\]
We have:
\[
    \bar{d} = \dfrac{1}{n}\sum_{i=1}^{n}(c_i - c_i^{'})
\] and 
\[ 
s_d = \sqrt{\dfrac{1}{n-1}\sum_{i=1}^{n}(d_i - \bar{d})^{2}}
\]the 
corresponding t-value is
\[
    t = \dfrac{\bar{d}}{\frac{s_d}{\sqrt{n}}}
\].

After calculating this t-value, we can calculate a probability $p$ (following the setting of T-test), which represents the probability of mis-rejecting the null hypothesis. If $p<0.05$, we can confidently reject null hypothesis and choose the alternative hypothesis, which means the model is statistically significantly more confident when answering the original questions and this provides evidence for potential contamination. 

In short, if the calculated $p < 0.05$, we say the model $M$ is contaminated on benchmark $D$, otherwise we say there is no statistically significant evidence of contamination. 

The whole process of our method is shown in Algorithm \ref{alg:1}.

\begin{algorithm}
    \caption{\OUR}            
    \begin{algorithmic}[1]
        \STATE Input benchmark 
        $D = \{(x_1, y_1), ..., (x_n, y_n)\}$ and model to test $M$, model used to rephrase $M_p$.
        \FOR{$i = 1, 2..., n$}
            \STATE $x^{'}_i \leftarrow M_{p}(x_i)$
            \STATE $c_i \leftarrow \mathcal{P}(Yes|x_i,M(x_i),M)$ 
            \STATE $c_{i}^{'} \leftarrow \mathcal{P}(Yes|x^{'}_i,M(x^{'}_i),M)$
        \ENDFOR
        \STATE $\bar{d}\leftarrow \dfrac{\sum_{i=1}^{n}c_{i}-c_{i}^{'}}{n}$
        \STATE $s_{d} \leftarrow \sqrt{\dfrac{1}{n-1}\sum_{i=1}^{n}(d_i - \bar{d})^{2}}$ 
        \STATE $t \leftarrow \dfrac{\bar{d}}{\frac{s_{d}}{\sqrt{n}}}$, Calculate $p$ according to $t$ and $n$
        \IF{p < 0.05 (Significant)}
            \STATE Return: $D$ is Contaminated
        \ELSE
            \STATE Return: $D$ is not Contaminated
        \ENDIF
        
    \end{algorithmic}
    \label{alg:1}
\end{algorithm} 
\section{Experiments}
\label{sec:5.1}


\begin{table*}[htbp]
    \small
    \centering  
    \setlength{\tabcolsep}{18pt}
    \begin{tabular}{cccc@{\extracolsep{4pt}}}  
    \toprule  
        Model  & Method & Trained Data & Untrained Data \\  
    \midrule[0.5pt]
    \multirow{3}{*}{\makecell[c]{Llama \\ (Contaminated)}} & Guided-Prompting & \underline{0.99} & 0.62 \\  
     ~ & PaCoST(simplified) & \underline{0.94} & 0.99 \\  
    ~ & PaCoST(ours) & \textbf{6e-8} & 0.92\\  
    \midrule 
    \multirow{3}{*}{\makecell[c]{Mistral \\ (Contaminated)}} & Guided-Prompting & \underline{0.99} & 0.99 \\  
    ~ & PaCoST(simplified) & \textbf{0.02} & 0.36 \\  
    ~ & PaCoST(ours) & \textbf{2e-4} & 0.75 \\  
    \midrule
    \multirow{3}{*}{\makecell[c]{Llama \\ (Original)}} & Guided-Prompting & \underline{\textbf{1e-10}} & \underline{\textbf{1e-9}}\\  
     ~ & PaCoST(simplified) & 0.78 & 0.87 \\  
     ~ & PaCoST(ours) & 0.12 & 0.92 \\ 
     \midrule
     \multirow{3}{*}{\makecell[c]{Mistral \\ (Original)}} & Guided-Prompting & \underline{\textbf{7e-5}} & \underline{\textbf{1e-3}} \\  
     ~ & PaCoST(simplified) & 0.18 & 0.46 \\  
     ~ & PaCoST(ours) & 0.46 & 0.72\\  
    \bottomrule  
    \end{tabular}  
    \caption{Main results of intentional contamination. The values are p-value of the methods, where $p<0.05$ represents statistically significant and probably contaminated and $p\geq 0.05$ represents un-contaminated. The bold p-values represents significant results. The underlined \underline{values} represent false positive or false negative results.}  
    \label{tab:main-results}  
\end{table*}

\subsection{Intentional Contamination Experiments}
First, to validate the effectiveness of our method, we conduct intentional contamination experiments. 

\paragraph{Experiment Settings}
For these experiments, we select Mistral-7B-Instruct-v0.2\citep{jiang2023mistral} and Llama-2-7B-Chat\citep{touvron2023llama2} as the target models and utilize a newly released dataset WMDP \cite{li2024wmdp} for intentional contamination. This dataset, including 3,668 multiple-choice questions about knowledge in biology, chemistry and cyber, is released in May 2024, ensuring that the selected models have not been contaminated on this data before. 

We conduct supervised fine-tuning (following the second contamination type) on two models. Though there are two contamination types as we introduced in Section \ref{sec:prop}, we mainly conduct intentional contamination experiments following the second contamination type because it is less discussed and somehow more difficult to detect because it has less trained parts. It is worth mentioning that our method works properly under the first contamination type, as is shown in Appendix \ref{sec:app-2}.

We sample 1000 samples from biology split from the WMDP dataset to produce contaminated versions of Llama and Mistral. 400 samples are sampled from the remaining data in the WMDP dataset to form "clean" (untrained) data. The choice of number of samples are just for simplicity and does not affect the final results as we will show later.

For baseline comparisons, given the limited availability of benchmark-level contamination detection methods, we selected Guided-Prompting \citep{golchin2023time} as our baseline. Since Guided-Prompting also utilizes p-values as an indicator of contamination, this allows for a fair comparison between our method and theirs. 

We also compare the performance of our method with a simplified version that directly uses ground truth answer to calculate confidence instead of the model's generated reponse. Details of the simplified version will be discussed in Appendix \ref{sec:app-simp}.

Additionally, we conduct experiments to evaluate the performance of DCQ \citep{golchin2023data} and Min-k\% Prob \citep{shi2024detecting} and find that they do not perform well for detecting benchmark contamination. A detailed discussion of these findings can be found in Appendix \ref{sec:app-comp}.

\paragraph{Results and Analysis}
 The results are presented in Table \ref{tab:main-results}. Our method successfully identifies contaminated datasets in contaminated models, demonstrated by significant results on trained data in these models. Importantly, our method avoids false positives, as it does not return significant results on uncontaminated datasets, even when applied to contaminated models. For original models, which are free from contamination, all results are insignificant. These findings underscore the effectiveness of our method in accurately detecting data contamination.

In contrast, Guided-Prompting fails to identify contamination in contaminated models, likely because the instruction part was not included in the training parts, preventing Guided-Prompting from replicating the original data accurately. Similarly, the simplified version of our method performs much better than Guided-Prompting, but it still suffers from false negative problems. These comparisons further reveal the effectiveness of our method. Some detailed discussions about this result can be found in Appendix \ref{sec:app-simp}.

\paragraph{Stability under Different Number of Samples}

Different datasets vary in the amount of data they contain, and for very large datasets, it is more practical to sample a subset for contamination detection. Therefore, it is crucial to validate that our method performs well with varying sample sizes. To test this, we conducted experiments under the same settings as above but with different numbers of samples. The results are presented in Table \ref{tab:stab-number}.

\begin{table}[htbp]
    \centering
    \small
    \begin{tabular}{cccc}
    \toprule
        Data & \makecell[c]{\#Sample} & \makecell[c]{Llama / Mistral  \\
        (Contaminated)} & \makecell[c]{Llama / Mistral\\ (Original)} \\
        \midrule
        \multirow{3}{*}{\makecell[c]{Trained \\Data}} & 1000 & \textbf{6e-8} / \textbf{2e-4} & 0.12 / 0.46  \\
        ~ & 500 & \textbf{1e-5} / \textbf{7e-8} & 0.41 / 0.55 \\
        ~ & 100 & \textbf{0.02} / \textbf{1e-3} & 0.81 / 0.38 \\
        \midrule
        \multirow{3}{*}{\makecell[c]{Untrained\\Data}} & 400 &  0.92 / 0.75 & 0.92 / 0.72\\
        ~ & 200 & 0.54 / 0.84  & 0.83 / 0.56\\
        ~ & 100 & 0.88 / 0.62 & 0.59 / 0.27\\
        \bottomrule
    \end{tabular}
    \caption{p-value of different number of samples. The significant results are in bold.}
    \label{tab:stab-number}
\end{table}

As indicated by the results, our method works properly with sample sizes ranging from 100 to 1000, without generating any false positives or false negatives. This demonstrates the stability of our method across different sample sizes and highlights that it only requires a subset of the dataset to effectively detect contamination, thereby reducing the cost of processing entire datasets.

We do not discuss samples with fewer than 100 instances for two reasons. First, because our method relies on statistical analysis, a small sample size can introduce significant randomness, which could interfere with accurate contamination detection. Second, datasets with fewer than 100 samples are rare, making the analysis of such scenarios less relevant and meaningful.

We also conducted additional studies to assess the behavior of our method under various conditions. We demonstrated that our method maintains stable performance when using different rephrase models $M_p$. It is also robust to reasonable randomness, as it delivers consistent performance under different random seeds. Furthermore, our method effectively handles various types of contamination. These findings collectively highlight the superiority of our method. Detailed discussions can be found in Appendix \ref{sec:app-2}.

\begin{table*}[ht]
\centering
\setlength{\tabcolsep}{4mm} 
\footnotesize
\begin{tabular}{lcccccc}
\toprule
\textbf{Model} & \textbf{Arc-c} & \textbf{Arc-e} & \textbf{MMLU} & \textbf{HellaS} & \textbf{WinoG} & \textbf{T-QA} \\
\midrule

Model I & 0.46 & 0.53 & \textbf{2e-3} & \textbf{3e-8} & 0.78  & 0.57 \\
Model II& 0.18 & 0.30 & \textbf{3e-7} & \textbf{7e-3} & 0.59 & 0.82 \\
Model III & \textbf{1e-3} & \textbf{1e-3}  & 0.30 & 0.37 & \textbf{2e-3} & \textbf{0.04} \\ 
Model IV &\textbf{0.02} & 0.28 & 0.09 & 0.59 & 0.25 &  \textbf{2e-3}\\ 
Model V &  \textbf{4e-4}& \textbf{7e-4} & 0.71 & 0.63 & 0.10 & 0.20 \\ 
Model VI & \textbf{1e-3} & \textbf{0.01} & \textbf{0.02} & 0.15 & \textbf{3e-8} &  0.24 \\
Model VII & 0.11 & \textbf{5e-3} & 0.73 & \textbf{0.03} & 0.11 &  0.82 \\ 
Model VIII & 0.09 & \textbf{0.04} & \textbf{0.02} & 0.10 & 0.26 &  0.44 \\ 
Model IX & 0.44 & \textbf{0.02} & 0.54 & \textbf{3e-13}  & \textbf{4e-3} &  \textbf{2e-8}  \\ 
Model X & 0.95 & 0.38 & 0.17 & 0.61 & 0.65 &  0.46 \\ 
\bottomrule
\end{tabular}
\caption{p-values of open-source models on widely tested benchmarks. (HellaS: HellaSwag, WinoG: WinoGrande, T-QA: TruthfulQA)}
\label{tab:full_auroc_wiki_orig_extended}
\end{table*}

\subsection{Tests on Existing LLMs and Benchmarks}

After showing the feasibility of our proposed method, we apply it to a variety of existing popular LLMs and benchmarks to assess their contamination status. In this section, we introduce the tested benchmarks, models, and present the experimental results and discussions. Since some benchmarks are extremely large, we randomly sample 400 samples in each benchmark for detection.

\paragraph{Datasets}
We conduct benchmark contamination detection experiments on some popular benchmarks, including MMLU \cite{mmlu}, HellaSwag \citep{zellers2019hellaswag}, Arc-E, Arc-C \cite{arc}, TruthfulQA \cite{truthfulqa}, WinoGrande \cite{winogrande}.

\paragraph{Models} We select the following open-source LLMs for experiments: Llama-2-Chat (7B, 13B) \cite{touvron2023llama2}, Llama-3-Instruct (8B) \cite{llama3modelcard}, Mistral-Instruct (7B) \cite{jiang2023mistral}, Phi-3 (3.8B) \cite{abdin2024phi3}, Qwen1.5 (0.5B, 7B), Qwen2 (7B) \cite{qwen}, Yi (6B) \cite{ai2024yi}, DeepSeek (7B) \cite{deepseek-llm}.

\paragraph{Evaluation Results}
We show the evaluation results in Table \ref{tab:full_auroc_wiki_orig_extended}. To avoid potential harmful effects, we do not show the names of the models in the results and represent them as Model I to Model X. Some observations can be drawn from the results. 

First of all, all benchmarks are suspected contaminated more or less on different models. Some benchmarks, like Arc-e, is suspected severely contaminated. Other benchmarks are are also suspected contaminated and we do not find a benchmark that is "clean" on all models.

Secondly, almost all models are suspected contaminated more or less on different benchmarks. Some models, like Model VI, Model IX, are suspected contaminated on 4 benchmarks out of 6 we tested. Other models are also suspected contaminated on 2 or 3 benchmarks out of 6 we tested. Model X is perhaps the "cleanest" model as we do not find significant evidence of contamination.

\subsection{Discussion}
This result further underscores the urgency of addressing the benchmark contamination problem in LLM evaluation. As evidenced, almost all models and benchmarks exhibit varying degrees of suspected contamination. This contamination undermines the trustworthiness of evaluation results on popular benchmarks, posing significant challenges for both users and developers.

It is important to note that we do not intend to accuse any LLM provider of intentional contamination. As previously discussed, given the vast amount of data required to train LLMs, excluding or even simply detecting benchmark data within training datasets is an exceedingly difficult task. We must acknowledge that benchmark contamination may be inevitable due to these constraints.

Instead, we would like to propose two key insights. First, detecting benchmark contamination is crucial because it allows us to assess whether evaluation results are trustworthy. While contamination does not inherently imply that a model is ineffective, recognizing its presence can prompt us to seek alternative evaluation metrics. This ensures that we are not misled by artificially high scores, and helps maintain the integrity and reliability of model evaluations.

Secondly, using specific benchmarks for evaluation may not be suitable. As our findings reveal, all benchmarks are suspected contaminated to some degree. As soon as a new benchmark is made public, it quickly becomes susceptible to contamination because LLMs require large-scale, high-quality data for training, and benchmarks naturally fit this criterion. However, if a benchmark is not released publicly, its quality and the evaluation results derived from it cannot be fully trusted, leading to a dilemma.

 Therefore, we advocate for a new LLM evaluation approach that does not rely on static benchmarks but rather on flexible and dynamic data sources. For instance, evaluating LLMs based on user feedback data, could provide a dynamic and resilient measure of model performance. Further, quantitative LLM evaluation can also be made public - everyone can build his own benchmark for evaluation. If the results of this large-scale benchmarks could be combined, the evaluation of LLMs will be more trustworthy and comprehensive.



\section{Conclusion}
In this work, we introduce the issue of benchmark contamination in LLMs and propose several essential criteria that an effective benchmark contamination detection method should meet. We highlight that all existing detection methods fall short of satisfying all of these requirements. We then propose a benchmark contamination detection method named \OUR, which uses significantly higher confidence scores as an indicator of contamination. We conduct various experiments to demonstrate the effectiveness of our method. Additionally, we apply our method to popular LLMs and benchmarks and reveal a significant problem of benchmark contamination across almost all benchmarks and LLMs we examined.

\section*{Limitations}
Our method focuses on detecting benchmark-level contamination and is not suitable for identifying instance-level contamination. Additionally, our method involves multiple interactions with the LLM, including one for paraphrasing, two for answer generation, and two for confidence estimation. This can result in lower efficiency compared to other approaches.

Moreover, our method requires access to the probability distribution for confidence estimation, which is not available in black-box LLMs. As a result, our approach cannot be used to detect benchmark contamination in black-box LLMs where internal outputs like probability distributions are not accessible.

\section*{Ethics Statement}
We honestly report the p-values for various open-source LLMs and benchmarks without any alteration to enhance or detract from the results. The intentionally contaminated checkpoints used in our research are for academic purposes only and will not be released because WMDP is a "dangerous" dataset that should be forgotten instead of memorized by models. The aim of this work is to highlight and address the issue of benchmark contamination, not to promote contamination or criticize any parties involved. We deeply respect the contributions of LLM and benchmark providers and believe that the problem of benchmark contamination will be effectively addressed in due course. ChatGPT is used only to assist writing.

\section*{Acknowledgement}
This work was supported by Beijing Science and Technology Program (Z231100007423011), National Science Foundation of China (No. 62161160339) and Key Laboratory of Science, Technology and Standard in Press Industry (Key Laboratory of Intelligent Press Media Technology). We appreciate the anonymous reviewers for their helpful comments. Xiaojun Wan is the corresponding author.

\bibliography{acl_latex}

\begin{thebibliography}{56}
\providecommand{\natexlab}[1]{#1}

\bibitem[{Abdin et~al.(2024)Abdin, Jacobs, Awan, Aneja, Awadallah, Awadalla, Bach, Bahree, Bakhtiari, Bao, Behl, Benhaim, Bilenko, Bjorck, Bubeck, Cai, Cai, Mendes, Chen, Chaudhary, Chen, Chen, Chen, Chen, Chopra, Dai, Giorno, de~Rosa, Dixon, Eldan, Fragoso, Iter, Gao, Gao, Gao, Garg, Goswami, Gunasekar, Haider, Hao, Hewett, Huynh, Javaheripi, Jin, Kauffmann, Karampatziakis, Kim, Khademi, Kurilenko, Lee, Lee, Li, Li, Liang, Liden, Liu, Liu, Liu, Lin, Lin, Luo, Madan, Mazzola, Mitra, Modi, Nguyen, Norick, Patra, Perez-Becker, Portet, Pryzant, Qin, Radmilac, Rosset, Roy, Ruwase, Saarikivi, Saied, Salim, Santacroce, Shah, Shang, Sharma, Shukla, Song, Tanaka, Tupini, Wang, Wang, Wang, Wang, Ward, Wang, Witte, Wu, Wyatt, Xiao, Xu, Xu, Xu, Yadav, Yang, Yang, Yang, Yang, Yu, Yuan, Zhang, Zhang, Zhang, Zhang, Zhang, Zhang, Zhang, and Zhou}]{abdin2024phi3}
Marah Abdin, Sam~Ade Jacobs, Ammar~Ahmad Awan, Jyoti Aneja, Ahmed Awadallah, Hany Awadalla, Nguyen Bach, Amit Bahree, Arash Bakhtiari, Jianmin Bao, Harkirat Behl, Alon Benhaim, Misha Bilenko, Johan Bjorck, Sébastien Bubeck, Qin Cai, Martin Cai, Caio César~Teodoro Mendes, Weizhu Chen, Vishrav Chaudhary, Dong Chen, Dongdong Chen, Yen-Chun Chen, Yi-Ling Chen, Parul Chopra, Xiyang Dai, Allie~Del Giorno, Gustavo de~Rosa, Matthew Dixon, Ronen Eldan, Victor Fragoso, Dan Iter, Mei Gao, Min Gao, Jianfeng Gao, Amit Garg, Abhishek Goswami, Suriya Gunasekar, Emman Haider, Junheng Hao, Russell~J. Hewett, Jamie Huynh, Mojan Javaheripi, Xin Jin, Piero Kauffmann, Nikos Karampatziakis, Dongwoo Kim, Mahoud Khademi, Lev Kurilenko, James~R. Lee, Yin~Tat Lee, Yuanzhi Li, Yunsheng Li, Chen Liang, Lars Liden, Ce~Liu, Mengchen Liu, Weishung Liu, Eric Lin, Zeqi Lin, Chong Luo, Piyush Madan, Matt Mazzola, Arindam Mitra, Hardik Modi, Anh Nguyen, Brandon Norick, Barun Patra, Daniel Perez-Becker, Thomas Portet, Reid Pryzant, Heyang
  Qin, Marko Radmilac, Corby Rosset, Sambudha Roy, Olatunji Ruwase, Olli Saarikivi, Amin Saied, Adil Salim, Michael Santacroce, Shital Shah, Ning Shang, Hiteshi Sharma, Swadheen Shukla, Xia Song, Masahiro Tanaka, Andrea Tupini, Xin Wang, Lijuan Wang, Chunyu Wang, Yu~Wang, Rachel Ward, Guanhua Wang, Philipp Witte, Haiping Wu, Michael Wyatt, Bin Xiao, Can Xu, Jiahang Xu, Weijian Xu, Sonali Yadav, Fan Yang, Jianwei Yang, Ziyi Yang, Yifan Yang, Donghan Yu, Lu~Yuan, Chengruidong Zhang, Cyril Zhang, Jianwen Zhang, Li~Lyna Zhang, Yi~Zhang, Yue Zhang, Yunan Zhang, and Xiren Zhou. 2024.
\newblock \href {https://arxiv.org/abs/2404.14219} {Phi-3 technical report: A highly capable language model locally on your phone}.
\newblock \emph{Preprint}, arXiv:2404.14219.

\bibitem[{AI et~al.(2024)AI, :, Young, Chen, Li, Huang, Zhang, Zhang, Li, Zhu, Chen, Chang, Yu, Liu, Liu, Yue, Yang, Yang, Yu, Xie, Huang, Hu, Ren, Niu, Nie, Xu, Liu, Wang, Cai, Gu, Liu, and Dai}]{ai2024yi}
01. AI, :, Alex Young, Bei Chen, Chao Li, Chengen Huang, Ge~Zhang, Guanwei Zhang, Heng Li, Jiangcheng Zhu, Jianqun Chen, Jing Chang, Kaidong Yu, Peng Liu, Qiang Liu, Shawn Yue, Senbin Yang, Shiming Yang, Tao Yu, Wen Xie, Wenhao Huang, Xiaohui Hu, Xiaoyi Ren, Xinyao Niu, Pengcheng Nie, Yuchi Xu, Yudong Liu, Yue Wang, Yuxuan Cai, Zhenyu Gu, Zhiyuan Liu, and Zonghong Dai. 2024.
\newblock \href {https://arxiv.org/abs/2403.04652} {Yi: Open foundation models by 01.ai}.
\newblock \emph{Preprint}, arXiv:2403.04652.

\bibitem[{AI@Meta(2024)}]{llama3modelcard}
AI@Meta. 2024.
\newblock \href {https://github.com/meta-llama/llama3/blob/main/MODEL_CARD.md} {Llama 3 model card}.

\bibitem[{Bai et~al.(2023)Bai, Bai, Chu, Cui, Dang, Deng, Fan, Ge, Han, Huang, Hui, Ji, Li, Lin, Lin, Liu, Liu, Lu, Lu, Ma, Men, Ren, Ren, Tan, Tan, Tu, Wang, Wang, Wang, Wu, Xu, Xu, Yang, Yang, Yang, Yang, Yao, Yu, Yuan, Yuan, Zhang, Zhang, Zhang, Zhang, Zhou, Zhou, Zhou, and Zhu}]{qwen}
Jinze Bai, Shuai Bai, Yunfei Chu, Zeyu Cui, Kai Dang, Xiaodong Deng, Yang Fan, Wenbin Ge, Yu~Han, Fei Huang, Binyuan Hui, Luo Ji, Mei Li, Junyang Lin, Runji Lin, Dayiheng Liu, Gao Liu, Chengqiang Lu, Keming Lu, Jianxin Ma, Rui Men, Xingzhang Ren, Xuancheng Ren, Chuanqi Tan, Sinan Tan, Jianhong Tu, Peng Wang, Shijie Wang, Wei Wang, Shengguang Wu, Benfeng Xu, Jin Xu, An~Yang, Hao Yang, Jian Yang, Shusheng Yang, Yang Yao, Bowen Yu, Hongyi Yuan, Zheng Yuan, Jianwei Zhang, Xingxuan Zhang, Yichang Zhang, Zhenru Zhang, Chang Zhou, Jingren Zhou, Xiaohuan Zhou, and Tianhang Zhu. 2023.
\newblock Qwen technical report.
\newblock \emph{arXiv preprint arXiv:2309.16609}.

\bibitem[{Brown et~al.(2020{\natexlab{a}})Brown, Mann, Ryder, Subbiah, Kaplan, Dhariwal, Neelakantan, Shyam, Sastry, Askell et~al.}]{brown2020language}
Tom Brown, Benjamin Mann, Nick Ryder, Melanie Subbiah, Jared~D Kaplan, Prafulla Dhariwal, Arvind Neelakantan, Pranav Shyam, Girish Sastry, Amanda Askell, et~al. 2020{\natexlab{a}}.
\newblock Language models are few-shot learners.
\newblock \emph{Advances in neural information processing systems}, 33:1877--1901.

\bibitem[{Brown et~al.(2020{\natexlab{b}})Brown, Mann, Ryder, Subbiah, Kaplan, Dhariwal, Neelakantan, Shyam, Sastry, Askell, Agarwal, Herbert-Voss, Krueger, Henighan, Child, Ramesh, Ziegler, Wu, Winter, Hesse, Chen, Sigler, Litwin, Gray, Chess, Clark, Berner, McCandlish, Radford, Sutskever, and Amodei}]{gpt3}
Tom~B. Brown, Benjamin Mann, Nick Ryder, Melanie Subbiah, Jared Kaplan, Prafulla Dhariwal, Arvind Neelakantan, Pranav Shyam, Girish Sastry, Amanda Askell, Sandhini Agarwal, Ariel Herbert-Voss, Gretchen Krueger, Tom Henighan, Rewon Child, Aditya Ramesh, Daniel~M. Ziegler, Jeffrey Wu, Clemens Winter, Christopher Hesse, Mark Chen, Eric Sigler, Mateusz Litwin, Scott Gray, Benjamin Chess, Jack Clark, Christopher Berner, Sam McCandlish, Alec Radford, Ilya Sutskever, and Dario Amodei. 2020{\natexlab{b}}.
\newblock Language models are few-shot learners.
\newblock In \emph{Proceedings of the 34th International Conference on Neural Information Processing Systems}, NIPS '20, Red Hook, NY, USA. Curran Associates Inc.

\bibitem[{Carlini et~al.(2021{\natexlab{a}})Carlini, Tramer, Wallace, Jagielski, Herbert-Voss, Lee, Roberts, Brown, Song, Erlingsson, Oprea, and Raffel}]{carlini2021extracting}
Nicholas Carlini, Florian Tramer, Eric Wallace, Matthew Jagielski, Ariel Herbert-Voss, Katherine Lee, Adam Roberts, Tom Brown, Dawn Song, Ulfar Erlingsson, Alina Oprea, and Colin Raffel. 2021{\natexlab{a}}.
\newblock \href {https://arxiv.org/abs/2012.07805} {Extracting training data from large language models}.
\newblock \emph{Preprint}, arXiv:2012.07805.

\bibitem[{Carlini et~al.(2021{\natexlab{b}})Carlini, Tramer, Wallace, Jagielski, Herbert-Voss, Lee, Roberts, Brown, Song, Erlingsson, and {others}}]{carlini_extracting_2021}
Nicholas Carlini, Florian Tramer, Eric Wallace, Matthew Jagielski, Ariel Herbert-Voss, Katherine Lee, Adam Roberts, Tom Brown, Dawn Song, Ulfar Erlingsson, and {others}. 2021{\natexlab{b}}.
\newblock Extracting training data from large language models.
\newblock In \emph{30th {USENIX} Security Symposium ({USENIX} Security 21)}, pages 2633--2650.

\bibitem[{Chen and Mueller(2023)}]{chen2023quantifying}
Jiuhai Chen and Jonas Mueller. 2023.
\newblock Quantifying uncertainty in answers from any language model via intrinsic and extrinsic confidence assessment.
\newblock \emph{arXiv preprint arXiv:2308.16175}.

\bibitem[{Chowdhery et~al.(2022)Chowdhery, Narang, Devlin, Bosma, Mishra, Roberts, Barham, Chung, Sutton, Gehrmann et~al.}]{chowdhery2022palm}
Aakanksha Chowdhery, Sharan Narang, Jacob Devlin, Maarten Bosma, Gaurav Mishra, Adam Roberts, Paul Barham, Hyung~Won Chung, Charles Sutton, Sebastian Gehrmann, et~al. 2022.
\newblock Palm: Scaling language modeling with pathways.
\newblock \emph{arXiv preprint arXiv:2204.02311}.

\bibitem[{Chowdhery et~al.(2023)Chowdhery, Narang, Devlin, Bosma, Mishra, Roberts, Barham, Chung, Sutton, Gehrmann et~al.}]{chowdhery2023palm}
Aakanksha Chowdhery, Sharan Narang, Jacob Devlin, Maarten Bosma, Gaurav Mishra, Adam Roberts, Paul Barham, Hyung~Won Chung, Charles Sutton, Sebastian Gehrmann, et~al. 2023.
\newblock Palm: Scaling language modeling with pathways.
\newblock \emph{Journal of Machine Learning Research}, 24(240):1--113.

\bibitem[{Clark et~al.(2018)Clark, Cowhey, Etzioni, Khot, Sabharwal, Schoenick, and Tafjord}]{arc}
Peter Clark, Isaac Cowhey, Oren Etzioni, Tushar Khot, Ashish Sabharwal, Carissa Schoenick, and Oyvind Tafjord. 2018.
\newblock Think you have solved question answering? try arc, the ai2 reasoning challenge.
\newblock \emph{arXiv preprint arXiv:1803.05457}.

\bibitem[{Cummings et~al.(2023)Cummings, Desfontaines, Evans, Geambasu, Jagielski, Huang, Kairouz, Kamath, Oh, Ohrimenko, and {others}}]{cummings_challenges_2023}
Rachel Cummings, Damien Desfontaines, David Evans, Roxana Geambasu, Matthew Jagielski, Yangsibo Huang, Peter Kairouz, Gautam Kamath, Sewoong Oh, Olga Ohrimenko, and {others}. 2023.
\newblock Challenges towards the next frontier in privacy.

\bibitem[{DeepSeek-AI(2024)}]{deepseek-llm}
DeepSeek-AI. 2024.
\newblock \href {https://github.com/deepseek-ai/DeepSeek-LLM} {Deepseek llm: Scaling open-source language models with longtermism}.
\newblock \emph{arXiv preprint arXiv:2401.02954}.

\bibitem[{Dong et~al.(2024)Dong, Jiang, Liu, Jin, and Li}]{dong2024generalization}
Yihong Dong, Xue Jiang, Huanyu Liu, Zhi Jin, and Ge~Li. 2024.
\newblock Generalization or memorization: Data contamination and trustworthy evaluation for large language models.
\newblock \emph{arXiv preprint arXiv:2402.15938}.

\bibitem[{Golchin and Surdeanu(2023{\natexlab{a}})}]{golchin2023data}
Shahriar Golchin and Mihai Surdeanu. 2023{\natexlab{a}}.
\newblock Data contamination quiz: A tool to detect and estimate contamination in large language models.
\newblock \emph{arXiv preprint arXiv:2311.06233}.

\bibitem[{Golchin and Surdeanu(2023{\natexlab{b}})}]{golchin2023time}
Shahriar Golchin and Mihai Surdeanu. 2023{\natexlab{b}}.
\newblock Time travel in llms: Tracing data contamination in large language models.
\newblock \emph{arXiv preprint arXiv:2308.08493}.

\bibitem[{Gupta et~al.(2022)Gupta, Huang, Zhong, Gao, Li, and Chen}]{gupta_recovering_2022}
Samyak Gupta, Yangsibo Huang, Zexuan Zhong, Tianyu Gao, Kai Li, and Danqi Chen. 2022.
\newblock Recovering private text in federated learning of language models.
\newblock 35:8130--8143.

\bibitem[{Hendrycks et~al.(2020)Hendrycks, Burns, Basart, Zou, Mazeika, Song, and Steinhardt}]{mmlu}
Dan Hendrycks, Collin Burns, Steven Basart, Andy Zou, Mantas Mazeika, Dawn Song, and Jacob Steinhardt. 2020.
\newblock Measuring massive multitask language understanding.
\newblock \emph{arXiv preprint arXiv:2009.03300}.

\bibitem[{Jagielski et~al.(2020)Jagielski, Ullman, and Oprea}]{jagielski_auditing_2020}
Matthew Jagielski, Jonathan Ullman, and Alina Oprea. 2020.
\newblock Auditing differentially private machine learning: How private is private sgd?
\newblock 33:22205--22216.

\bibitem[{Jayaraman and Evans(2019)}]{jayaraman_evaluating_2019}
Bargav Jayaraman and David Evans. 2019.
\newblock Evaluating differentially private machine learning in practice.
\newblock In \emph{28th {USENIX} Security Symposium ({USENIX} Security 19)}, pages 1895--1912.

\bibitem[{Jiang et~al.(2023{\natexlab{a}})Jiang, Sablayrolles, Mensch, Bamford, Chaplot, Casas, Bressand, Lengyel, Lample, Saulnier et~al.}]{jiang2023mistral}
Albert~Q Jiang, Alexandre Sablayrolles, Arthur Mensch, Chris Bamford, Devendra~Singh Chaplot, Diego de~las Casas, Florian Bressand, Gianna Lengyel, Guillaume Lample, Lucile Saulnier, et~al. 2023{\natexlab{a}}.
\newblock Mistral 7b.
\newblock \emph{arXiv preprint arXiv:2310.06825}.

\bibitem[{Jiang et~al.(2023{\natexlab{b}})Jiang, Ruan, Huang, Liao, Pitis, Grosse, and Ba}]{jiang2023calibrating}
Mingjian Jiang, Yangjun Ruan, Sicong Huang, Saifei Liao, Silviu Pitis, Roger~Baker Grosse, and Jimmy Ba. 2023{\natexlab{b}}.
\newblock Calibrating language models via augmented prompt ensembles.

\bibitem[{Kadavath et~al.(2022)Kadavath, Conerly, Askell, Henighan, Drain, Perez, Schiefer, Hatfield-Dodds, DasSarma, Tran-Johnson et~al.}]{kadavath2022language}
Saurav Kadavath, Tom Conerly, Amanda Askell, Tom Henighan, Dawn Drain, Ethan Perez, Nicholas Schiefer, Zac Hatfield-Dodds, Nova DasSarma, Eli Tran-Johnson, et~al. 2022.
\newblock Language models (mostly) know what they know.
\newblock \emph{arXiv preprint arXiv:2207.05221}.

\bibitem[{Kuhn et~al.(2023)Kuhn, Gal, and Farquhar}]{kuhn2023semantic}
Lorenz Kuhn, Yarin Gal, and Sebastian Farquhar. 2023.
\newblock Semantic uncertainty: Linguistic invariances for uncertainty estimation in natural language generation.
\newblock \emph{arXiv preprint arXiv:2302.09664}.

\bibitem[{Li et~al.(2024)Li, Pan, Gopal, Yue, Berrios, Gatti, Li, Dombrowski, Goel, Phan, Mukobi, Helm-Burger, Lababidi, Justen, Liu, Chen, Barrass, Zhang, Zhu, Tamirisa, Bharathi, Khoja, Zhao, Herbert-Voss, Breuer, Marks, Patel, Zou, Mazeika, Wang, Oswal, Liu, Hunt, Tienken-Harder, Shih, Talley, Guan, Kaplan, Steneker, Campbell, Jokubaitis, Levinson, Wang, Qian, Karmakar, Basart, Fitz, Levine, Kumaraguru, Tupakula, Varadharajan, Shoshitaishvili, Ba, Esvelt, Wang, and Hendrycks}]{li2024wmdp}
Nathaniel Li, Alexander Pan, Anjali Gopal, Summer Yue, Daniel Berrios, Alice Gatti, Justin~D. Li, Ann-Kathrin Dombrowski, Shashwat Goel, Long Phan, Gabriel Mukobi, Nathan Helm-Burger, Rassin Lababidi, Lennart Justen, Andrew~B. Liu, Michael Chen, Isabelle Barrass, Oliver Zhang, Xiaoyuan Zhu, Rishub Tamirisa, Bhrugu Bharathi, Adam Khoja, Zhenqi Zhao, Ariel Herbert-Voss, Cort~B. Breuer, Samuel Marks, Oam Patel, Andy Zou, Mantas Mazeika, Zifan Wang, Palash Oswal, Weiran Liu, Adam~A. Hunt, Justin Tienken-Harder, Kevin~Y. Shih, Kemper Talley, John Guan, Russell Kaplan, Ian Steneker, David Campbell, Brad Jokubaitis, Alex Levinson, Jean Wang, William Qian, Kallol~Krishna Karmakar, Steven Basart, Stephen Fitz, Mindy Levine, Ponnurangam Kumaraguru, Uday Tupakula, Vijay Varadharajan, Yan Shoshitaishvili, Jimmy Ba, Kevin~M. Esvelt, Alexandr Wang, and Dan Hendrycks. 2024.
\newblock \href {https://arxiv.org/abs/2403.03218} {The wmdp benchmark: Measuring and reducing malicious use with unlearning}.
\newblock \emph{Preprint}, arXiv:2403.03218.

\bibitem[{Li(2023)}]{li2023estimating}
Yucheng Li. 2023.
\newblock Estimating contamination via perplexity: Quantifying memorisation in language model evaluation.
\newblock \emph{arXiv preprint arXiv:2309.10677}.

\bibitem[{Lin et~al.(2021)Lin, Hilton, and Evans}]{truthfulqa}
Stephanie Lin, Jacob Hilton, and Owain Evans. 2021.
\newblock Truthfulqa: Measuring how models mimic human falsehoods.
\newblock \emph{arXiv preprint arXiv:2109.07958}.

\bibitem[{Lin et~al.(2022)Lin, Hilton, and Evans}]{lin2022teaching}
Stephanie Lin, Jacob Hilton, and Owain Evans. 2022.
\newblock Teaching models to express their uncertainty in words.
\newblock \emph{arXiv preprint arXiv:2205.14334}.

\bibitem[{Lin et~al.(2023)Lin, Trivedi, and Sun}]{lin2023generating}
Zhen Lin, Shubhendu Trivedi, and Jimeng Sun. 2023.
\newblock Generating with confidence: Uncertainty quantification for black-box large language models.
\newblock \emph{arXiv preprint arXiv:2305.19187}.

\bibitem[{Mahloujifar et~al.(2021)Mahloujifar, Inan, Chase, Ghosh, and Hasegawa}]{mahloujifar_membership_2021}
Saeed Mahloujifar, Huseyin~A Inan, Melissa Chase, Esha Ghosh, and Marcello Hasegawa. 2021.
\newblock Membership inference on word embedding and beyond.

\bibitem[{Mattern et~al.(2023)Mattern, Mireshghallah, Jin, Schoelkopf, Sachan, and Berg-Kirkpatrick}]{mattern-etal-2023-membership}
Justus Mattern, Fatemehsadat Mireshghallah, Zhijing Jin, Bernhard Schoelkopf, Mrinmaya Sachan, and Taylor Berg-Kirkpatrick. 2023.
\newblock \href {https://doi.org/10.18653/v1/2023.findings-acl.719} {Membership inference attacks against language models via neighbourhood comparison}.
\newblock In \emph{Findings of the Association for Computational Linguistics: ACL 2023}, pages 11330--11343, Toronto, Canada. Association for Computational Linguistics.

\bibitem[{Mireshghallah et~al.(2022)Mireshghallah, Goyal, Uniyal, Berg-Kirkpatrick, and Shokri}]{mireshghallah_quantifying_2022}
Fatemehsadat Mireshghallah, Kartik Goyal, Archit Uniyal, Taylor Berg-Kirkpatrick, and Reza Shokri. 2022.
\newblock \href {https://doi.org/10.18653/v1/2022.emnlp-main.570} {Quantifying privacy risks of masked language models using membership inference attacks}.
\newblock In \emph{Proceedings of the 2022 Conference on Empirical Methods in Natural Language Processing}, pages 8332--8347. Association for Computational Linguistics.

\bibitem[{Nasr et~al.(2023)Nasr, Carlini, Hayase, Jagielski, Cooper, Ippolito, Choquette-Choo, Wallace, Tram{\`e}r, and Lee}]{nasr2023scalable}
Milad Nasr, Nicholas Carlini, Jonathan Hayase, Matthew Jagielski, A~Feder Cooper, Daphne Ippolito, Christopher~A Choquette-Choo, Eric Wallace, Florian Tram{\`e}r, and Katherine Lee. 2023.
\newblock Scalable extraction of training data from (production) language models.
\newblock \emph{arXiv preprint arXiv:2311.17035}.

\bibitem[{OpenAI(2023)}]{openai2023gpt4}
OpenAI. 2023.
\newblock \href {https://arxiv.org/abs/2303.08774} {Gpt-4 technical report}.
\newblock \emph{Preprint}, arXiv:2303.08774.

\bibitem[{Oren et~al.(2023)Oren, Meister, Chatterji, Ladhak, and Hashimoto}]{oren2023proving}
Yonatan Oren, Nicole Meister, Niladri Chatterji, Faisal Ladhak, and Tatsunori~B Hashimoto. 2023.
\newblock Proving test set contamination in black box language models.
\newblock \emph{arXiv preprint arXiv:2310.17623}.

\bibitem[{Radford et~al.(2019)Radford, Wu, Child, Luan, Amodei, Sutskever et~al.}]{radford2019language}
Alec Radford, Jeffrey Wu, Rewon Child, David Luan, Dario Amodei, Ilya Sutskever, et~al. 2019.
\newblock Language models are unsupervised multitask learners.
\newblock \emph{OpenAI blog}, 1(8):9.

\bibitem[{Sainz et~al.(2023{\natexlab{a}})Sainz, Campos, Garc{\'\i}a-Ferrero, Etxaniz, de~Lacalle, and Agirre}]{sainz2023nlp}
Oscar Sainz, Jon~Ander Campos, Iker Garc{\'\i}a-Ferrero, Julen Etxaniz, Oier~Lopez de~Lacalle, and Eneko Agirre. 2023{\natexlab{a}}.
\newblock Nlp evaluation in trouble: On the need to measure llm data contamination for each benchmark.
\newblock \emph{arXiv preprint arXiv:2310.18018}.

\bibitem[{Sainz et~al.(2023{\natexlab{b}})Sainz, Campos, García-Ferrero, Etxaniz, and Agirre}]{did-ChatGPT-cheat-on-your-test}
Oscar Sainz, Jon~Ander Campos, Iker García-Ferrero, Julen Etxaniz, and Eneko Agirre. 2023{\natexlab{b}}.
\newblock Did chatgpt cheat on your test?
\newblock \url{https://hitz-zentroa.github.io/lm-contamination/blog/}.
\newblock Accessed: 2023-07-06.

\bibitem[{Sakaguchi et~al.(2021)Sakaguchi, Bras, Bhagavatula, and Choi}]{winogrande}
Keisuke Sakaguchi, Ronan~Le Bras, Chandra Bhagavatula, and Yejin Choi. 2021.
\newblock Winogrande: An adversarial winograd schema challenge at scale.
\newblock \emph{Communications of the ACM}, 64(9):99--106.

\bibitem[{Shejwalkar et~al.(2021)Shejwalkar, Inan, Houmansadr, and Sim}]{shejwalkar_membership_2021}
Virat Shejwalkar, Huseyin~A. Inan, Amir Houmansadr, and Robert Sim. 2021.
\newblock \href {https://openreview.net/forum?id=74lwg5oxheC} {Membership inference attacks against {NLP} classification models}.
\newblock In \emph{{NeurIPS} 2021 Workshop Privacy in Machine Learning}.

\bibitem[{Shi et~al.(2024)Shi, Ajith, Xia, Huang, Liu, Blevins, Chen, and Zettlemoyer}]{shi2024detecting}
Weijia Shi, Anirudh Ajith, Mengzhou Xia, Yangsibo Huang, Daogao Liu, Terra Blevins, Danqi Chen, and Luke Zettlemoyer. 2024.
\newblock \href {https://openreview.net/forum?id=zWqr3MQuNs} {Detecting pretraining data from large language models}.
\newblock In \emph{The Twelfth International Conference on Learning Representations}.

\bibitem[{Shokri et~al.(2017)Shokri, Stronati, Song, and Shmatikov}]{shokri_membership_2017-1}
Reza Shokri, Marco Stronati, Congzheng Song, and Vitaly Shmatikov. 2017.
\newblock Membership inference attacks against machine learning models.
\newblock In \emph{2017 {IEEE} symposium on security and privacy ({SP})}, pages 3--18. {IEEE}.

\bibitem[{Song and Shmatikov(2019)}]{song_auditing_2019}
Congzheng Song and Vitaly Shmatikov. 2019.
\newblock Auditing data provenance in text-generation models.
\newblock In \emph{Proceedings of the 25th {ACM} {SIGKDD} International Conference on Knowledge Discovery \& Data Mining}, pages 196--206.

\bibitem[{Tian et~al.(2023)Tian, Mitchell, Zhou, Sharma, Rafailov, Yao, Finn, and Manning}]{tian2023just}
Katherine Tian, Eric Mitchell, Allan Zhou, Archit Sharma, Rafael Rafailov, Huaxiu Yao, Chelsea Finn, and Christopher~D Manning. 2023.
\newblock Just ask for calibration: Strategies for eliciting calibrated confidence scores from language models fine-tuned with human feedback.
\newblock \emph{arXiv preprint arXiv:2305.14975}.

\bibitem[{Touvron et~al.(2023)Touvron, Martin, Stone, Albert, Almahairi, Babaei, Bashlykov, Batra, Bhargava, Bhosale, Bikel, Blecher, Ferrer, Chen, Cucurull, Esiobu, Fernandes, Fu, Fu, Fuller, Gao, Goswami, Goyal, Hartshorn, Hosseini, Hou, Inan, Kardas, Kerkez, Khabsa, Kloumann, Korenev, Koura, Lachaux, Lavril, Lee, Liskovich, Lu, Mao, Martinet, Mihaylov, Mishra, Molybog, Nie, Poulton, Reizenstein, Rungta, Saladi, Schelten, Silva, Smith, Subramanian, Tan, Tang, Taylor, Williams, Kuan, Xu, Yan, Zarov, Zhang, Fan, Kambadur, Narang, Rodriguez, Stojnic, Edunov, and Scialom}]{touvron2023llama2}
Hugo Touvron, Louis Martin, Kevin Stone, Peter Albert, Amjad Almahairi, Yasmine Babaei, Nikolay Bashlykov, Soumya Batra, Prajjwal Bhargava, Shruti Bhosale, Dan Bikel, Lukas Blecher, Cristian~Canton Ferrer, Moya Chen, Guillem Cucurull, David Esiobu, Jude Fernandes, Jeremy Fu, Wenyin Fu, Brian Fuller, Cynthia Gao, Vedanuj Goswami, Naman Goyal, Anthony Hartshorn, Saghar Hosseini, Rui Hou, Hakan Inan, Marcin Kardas, Viktor Kerkez, Madian Khabsa, Isabel Kloumann, Artem Korenev, Punit~Singh Koura, Marie-Anne Lachaux, Thibaut Lavril, Jenya Lee, Diana Liskovich, Yinghai Lu, Yuning Mao, Xavier Martinet, Todor Mihaylov, Pushkar Mishra, Igor Molybog, Yixin Nie, Andrew Poulton, Jeremy Reizenstein, Rashi Rungta, Kalyan Saladi, Alan Schelten, Ruan Silva, Eric~Michael Smith, Ranjan Subramanian, Xiaoqing~Ellen Tan, Binh Tang, Ross Taylor, Adina Williams, Jian~Xiang Kuan, Puxin Xu, Zheng Yan, Iliyan Zarov, Yuchen Zhang, Angela Fan, Melanie Kambadur, Sharan Narang, Aurelien Rodriguez, Robert Stojnic, Sergey Edunov, and Thomas
  Scialom. 2023.
\newblock \href {https://arxiv.org/abs/2307.09288} {Llama 2: Open foundation and fine-tuned chat models}.
\newblock \emph{Preprint}, arXiv:2307.09288.

\bibitem[{Wang et~al.(2019)Wang, Pruksachatkun, Nangia, Singh, Michael, Hill, Levy, and Bowman}]{wang2019superglue}
Alex Wang, Yada Pruksachatkun, Nikita Nangia, Amanpreet Singh, Julian Michael, Felix Hill, Omer Levy, and Samuel~R Bowman. 2019.
\newblock Superglue: A stickier benchmark for general-purpose language understanding systems.
\newblock \emph{arXiv preprint arXiv:1905.00537}.

\bibitem[{Wang et~al.(2022)Wang, Wei, Schuurmans, Le, Chi, Narang, Chowdhery, and Zhou}]{wang2022self}
Xuezhi Wang, Jason Wei, Dale Schuurmans, Quoc Le, Ed~Chi, Sharan Narang, Aakanksha Chowdhery, and Denny Zhou. 2022.
\newblock Self-consistency improves chain of thought reasoning in language models.
\newblock \emph{arXiv preprint arXiv:2203.11171}.

\bibitem[{Wei et~al.(2022)Wei, Bosma, Zhao, Guu, Yu, Lester, Du, Dai, and Le}]{wei2022finetuned}
Jason Wei, Maarten Bosma, Vincent~Y. Zhao, Kelvin Guu, Adams~Wei Yu, Brian Lester, Nan Du, Andrew~M. Dai, and Quoc~V. Le. 2022.
\newblock \href {https://arxiv.org/abs/2109.01652} {Finetuned language models are zero-shot learners}.
\newblock \emph{Preprint}, arXiv:2109.01652.

\bibitem[{Xiong et~al.(2023)Xiong, Hu, Lu, Li, Fu, He, and Hooi}]{xiong2023can}
Miao Xiong, Zhiyuan Hu, Xinyang Lu, Yifei Li, Jie Fu, Junxian He, and Bryan Hooi. 2023.
\newblock Can llms express their uncertainty? an empirical evaluation of confidence elicitation in llms.
\newblock \emph{arXiv preprint arXiv:2306.13063}.

\bibitem[{Yang et~al.(2023)Yang, Chiang, Zheng, Gonzalez, and Stoica}]{yang2023rethinking}
Shuo Yang, Wei-Lin Chiang, Lianmin Zheng, Joseph~E Gonzalez, and Ion Stoica. 2023.
\newblock Rethinking benchmark and contamination for language models with rephrased samples.
\newblock \emph{arXiv preprint arXiv:2311.04850}.

\bibitem[{Yeom et~al.(2018)Yeom, Giacomelli, Fredrikson, and Jha}]{yeom_privacy_2018}
Samuel Yeom, Irene Giacomelli, Matt Fredrikson, and Somesh Jha. 2018.
\newblock \href {https://doi.org/10.1109/CSF.2018.00027} {Privacy risk in machine learning: Analyzing the connection to overfitting}.
\newblock In \emph{2018 {IEEE} 31st Computer Security Foundations Symposium ({CSF})}, pages 268--282.

\bibitem[{Zellers et~al.(2019)Zellers, Holtzman, Bisk, Farhadi, and Choi}]{zellers2019hellaswag}
Rowan Zellers, Ari Holtzman, Yonatan Bisk, Ali Farhadi, and Yejin Choi. 2019.
\newblock Hellaswag: Can a machine really finish your sentence?
\newblock In \emph{Proceedings of the 57th Annual Meeting of the Association for Computational Linguistics}.

\bibitem[{Zhang et~al.(2024)Zhang, Da, Lee, Robinson, Wu, Song, Zhao, Raja, Slack, Lyu, Hendryx, Kaplan, Lunati, and Yue}]{zhang2024careful}
Hugh Zhang, Jeff Da, Dean Lee, Vaughn Robinson, Catherine Wu, Will Song, Tiffany Zhao, Pranav Raja, Dylan Slack, Qin Lyu, Sean Hendryx, Russell Kaplan, Michele Lunati, and Summer Yue. 2024.
\newblock \href {https://arxiv.org/abs/2405.00332} {A careful examination of large language model performance on grade school arithmetic}.
\newblock \emph{Preprint}, arXiv:2405.00332.

\bibitem[{Zhang* et~al.(2020)Zhang*, Kishore*, Wu*, Weinberger, and Artzi}]{bert-score}
Tianyi Zhang*, Varsha Kishore*, Felix Wu*, Kilian~Q. Weinberger, and Yoav Artzi. 2020.
\newblock \href {https://openreview.net/forum?id=SkeHuCVFDr} {Bertscore: Evaluating text generation with bert}.
\newblock In \emph{International Conference on Learning Representations}.

\bibitem[{Zhao et~al.(2023)Zhao, Yan, Sun, Xing, Meng, Wang, Cheng, Ren, and Yin}]{zhao2023knowing}
Yukun Zhao, Lingyong Yan, Weiwei Sun, Guoliang Xing, Chong Meng, Shuaiqiang Wang, Zhicong Cheng, Zhaochun Ren, and Dawei Yin. 2023.
\newblock Knowing what llms do not know: A simple yet effective self-detection method.
\newblock \emph{arXiv preprint arXiv:2310.17918}.

\end{thebibliography}

\appendix


\section{Discussions about Intentional Contamination Experiment}
\label{sec:app-simp}

\paragraph{Details about Simplified Version of Our Method}
We briefly introduce the simplified version of our method. Recall that our method calculate confidence $c_i = \mathcal{P}(Yes|x_i, M(x_i), M)$ for a given instance $(x_i, y_i)$. But it is natural to question whether it is possible to use $\widetilde{c}_i = \mathcal{P}(Yes|x_i, y_i, M)$, that is, to directly calculate model's "confidence" towards the ground truth answer. So we design a simplified version of our method in Algorithm \ref{alg:2}.

\begin{algorithm}
    \caption{\OUR (Simplified)}            
    \begin{algorithmic}[1]
        \STATE Input benchmark \\
        $D = \{(x_1, y_1), ..., (x_n, y_n)\}$ and model to test $M$, model used to rephrase $M_p$.
        \FOR{$i = 1, 2..., n$}
            \STATE $x^{'} \leftarrow M_{p}(x_i)$
            \STATE $\widetilde{c}_i \leftarrow \mathcal{P}(Yes|x_i,y_i,M)$ 
            \STATE $\widetilde{c}_{i}^{'} \leftarrow \mathcal{P}(Yes|x^{'}_i,y_i,M)$
        \ENDFOR
        \STATE $\bar{d}\leftarrow \dfrac{\sum_{i=1}^{n}\widetilde{c}_{i}-\widetilde{c}_{i}^{'}}{n}$
        \STATE $s_{d} \leftarrow \sqrt{\dfrac{1}{n-1}\sum_{i=1}^{n}(d_i - \bar{d})^{2}}$ 
        \STATE $t \leftarrow \dfrac{\bar{d}}{\frac{s_{d}}{\sqrt{n}}}$, Calculate $p$ according to $t$ and $n$
        \IF{p < 0.05 (Significant)}
            \STATE Return: $D$ is Contaminated
        \ELSE
            \STATE Return: $D$ is not Contaminated
        \ENDIF
        
    \end{algorithmic}
    \label{alg:2}
\end{algorithm} 

\paragraph{Discussion about Guided-Prompting}
Surprisingly, we find that Guided-Prompting generates numerous false-positive results on uncontaminated models. Since the WMDP dataset was released after the model checkpoints were created, and given that WMDP was authored by human experts \citep{li2024wmdp}, it is highly unlikely that WMDP was initially contaminated. Even if it were initially contaminated, Guided-Prompting should have been able to detect this in the subsequently contaminated checkpoints, which it failed to do. This observation further supports our assertion that Guided-Prompting is unstable across different prompts. The significance indicated by Guided-Prompting may stem from this instability rather than from genuine contamination.

\paragraph{Discussion about Simplified Version}
The simplified version of our method works much better than Guided-Prompting, as it correctly identifies one contaminated case and all un-contaminated cases. However, it makes a false negative mistake on contaminated Llama, making it less effective compared with the original version of PaCoST. 

We would like to attribute this false negative to the same reason mentioned in \citet{yang2023rethinking}, which argues that contaminated models would bear similar high performance even on rephrased samples. Therefore, using ground-truth answer may result in contaminated model behaving similarly on original samples and rephrased samples, leading to false negative mistakes. In contrast, our focus is that model will be more confident when \textbf{answering the question} instead of \textbf{towards the correct answer}. As can be seen from results in Table \ref{tab:main-results}, this assumption is more accurate and works better. Though the simplified version works well under some circumstances, our whole PaCoST performs better. 
 
\section{Comparison with Other Methods}
\label{sec:app-comp}

There are also many methods aiming at detecting contamination that are worth discussing. We mainly discuss two of them: DCQ \citep{golchin2023data} and Min-k\% Prob \citep{shi2024detecting}. 

\paragraph{Discussion of DCQ}

DCQ is a replication-based method which posits that models can distinguish between data they have been trained on and similar data they have not encountered during training. This method employs a multiple-choice quiz to detect contamination. We apply this method in our experiments and reported the accuracy in Table \ref{tab:dcq}. 

\begin{table}[htbp]
    \centering
    \small
    \begin{tabular}{ccc}
    \toprule
        Model & \makecell[c]{Trained Data} & \makecell[c]{Untrained Data}\\ 
        \midrule
       \makecell[c]{Llama  (Cont.)}  & 0.5 & 0.39 \\
       \bottomrule
    \end{tabular}
    \caption{Accuracy of DCQ. Cont. represents contaminated.}
    \label{tab:dcq}
\end{table}

As evident from the results, the accuracy is even worse than random guessing—random guessing would yield an accuracy of approximately 0.5. We believe this outcome is due to the following reasons.

First of all, the contaminated Llama follows the second contamination type, where only the answer part, not the instruction part, is trained. However, DCQ requires the model to identify the exact instruction part from multiple choices, which is particularly challenging given that the instruction part was not part of the training. This mismatch likely contributes to the method's poor performance in our experiments.

Secondly, numerous studies have demonstrated that LLMs are highly sensitive to prompts, and the order of choices in a multiple-choice question can significantly influence the outcome. This sensitivity leads to considerable variability in the method's performance, making it unreliable. As a result, users cannot draw definitive conclusions from its results due to this inherent instability.

\paragraph{Discussion of Min-k\% Prob}
Min-k\% Prob focuses on the k\% tokens with the smallest probabilities and 
$k$ is set to 20 to achieve the best performance according to \citet{shi2024detecting}. This method has two problems. The first one is, traditional Min-k\% Prob also requires the instruction to be trained. However, we do can adapt this method to only work on the answer part of a piece of data. But for relatively short trained parts (like an answer), 20\% tokens are simply one or two tokens, which may introduce too much randomness. The second one is, it requires a pre-defined threshold to determine contamination, but this threshold is hard to choose.

We report the accuracy of Min-k\% Prob in Table \ref{tab:mink}. We select $k=20$ and threshold $\epsilon = 0.1$. Specifically, if and only if the average probability of the min-20\% tokens is larger than 0.1, we classify the instance as contaminated. We present the accuracy results for both the original Min-k\% Prob and our adapted version of Min-k\% Prob.

\begin{table}[htbp]
    \centering
    \small
    \begin{tabular}{c|ccc}
        \toprule
        Method & Model & Trained  & Untrained \\
        \midrule
        \multirow{2}{*}{\makecell[c]{Min-k\% Prob \\ (Original)}} & Llama(Cont.) & 0.02 & 0.97 \\ 
        ~ & Llama(Original) &  0.94 & 0.98 \\
        \midrule
        \multirow{2}{*}{\makecell[c]{Min-k\% Prob \\ (Adapted)}} & Llama(Cont.) & 0.86 & 0.6 \\
        ~ & Llama(Original) & 1.0 & 1.0 \\
        \bottomrule
    \end{tabular}
    \caption{Accuracy of Min-k\% Prob. Cont. represents contaminated.}
    \label{tab:mink}
\end{table}

There are several interesting observations based on the results. First, the original Min-k\% Prob fails to determine contamination in the contaminated model because the instruction part is not trained. This aligns with our previous discussion.

The adapted Min-k\% Prob performs much better on both datasets. However, we observe an interesting phenomenon: for uncontaminated Llama, the model tends to output a relatively long response, causing the answer itself to have a relatively small probability, which leads to high detection accuracy. For contaminated Llama, the model outputs a single choice as response, but the probability of this choice is very high (e.g., 0.99999) no matter it is correct or not.

As a result, the contamination detection accuracy essentially becomes the accuracy of question answering. For contaminated data, if the model correctly answers a question, it outputs a very high probability, leading Min-k\% Prob to classify it as contaminated. Similarly, for uncontaminated data, if the model correctly answers a question, it also outputs a very high probability, still causing Min-k\% Prob to classify it as contaminated. Thus, in this case, Min-k\% Prob is effectively detecting whether the question is correctly answered, rather than whether the question is contaminated.

This observation also highlights the problems of using answer probabilities as a confidence score or using perplexity to determine contamination. Simple probabilities are easily influenced by various factors, including formatting, leading to unreliable results.

\section{Discussions of Our Method}
\label{sec:app-2}

In this part, we would like to make some detailed discussions about our method to show that our method provides stable and trustworthy results. For simplicity, the following experiments are conducted on Llama only. 

\paragraph{Quality of Rephrasing}
\begin{table*}[htbp]
\centering
\small 
\renewcommand{\arraystretch}{1.2} 
\begin{tabular}{m{3cm} c c c c c c c c c c c c}
\toprule
  \multirow{2}{*}{Model} &
  \multicolumn{5}{c}{\makecell{\rule{0pt}{1ex}Trained Data}} & 
  \multicolumn{1}{c}{} &
  \multicolumn{5}{c}{\makecell{\rule{0pt}{1ex}Untrained Data}} \\ \cline{2-6} \cline{8-12} 
   &
  \makecell{\rule{0pt}{1ex}0} & 
  \makecell{\rule{0pt}{1ex}42} &
  \makecell{\rule{0pt}{1ex}302} &
  \makecell{\rule{0pt}{1ex}3407} &
  \makecell{\rule{0pt}{1ex}9056} &
   &
  \multicolumn{1}{c}{0} &
  \multicolumn{1}{c}{42} &
  \multicolumn{1}{c}{302} & 
  \multicolumn{1}{c}{3407} & 
  \multicolumn{1}{c}{9056} & \\ \midrule
  Llama (Contaminated) &
  \textbf{6e-4} &
  \textbf{4e-5} &
  \textbf{9e-6} &
    \textbf{4e-8}&
  \textbf{0.01} 
  &
  & 0.94
  & 0.96
  & 0.97 
  & 0.63 
  & 0.97
   \\
 
  Llama (Original) & 0.73
   & 0.98
   & 0.98
   & 0.83
   & 0.77
   &
   & 0.99
   & 0.92 
   & 0.99
   & 0.99
   & 0.99 \\ \bottomrule
\end{tabular}
\caption{p-value of different random seeds. The significant results are in bold.}
\label{tab:stab:rand}
\end{table*}
Though LLMs are known to handle various tasks effectively, it is still reasonable to question their proficiency at rephrasing. If the rephrasing model $M_p$ fails to correctly rephrase a question, the results of our method would become meaningless. Therefore, we aim to investigate the quality of rephrasing.

Since we primarily use Llama-2-Chat-7B for rephrasing, we focus on evaluating its rephrasing quality. We use the same dataset split mentioned in Section \ref{sec:5.1} and randomly sample 100 instances from each split to evaluate the quality of rephrasing. We use two evaluation methods: BERT-Score \citep{bert-score} and human study. We employ two human annotators to check whether each rephrasing result is correct (i.e., it does not change the original meaning and is not exactly the same as the original instance) and annotate each as 0 (incorrect) or 1 (correct). The results are shown in Table \ref{tab:para-qua}.

\begin{table}[htbp]
    \centering
    \small
    \begin{tabular}{c|cc}
    \toprule
        Data & BERT-Score & Human Evaluation \\
    \midrule
        Trained & 0.95 & 0.89 \\
        Untrained & 0.94 & 0.91 \\
    \bottomrule
    \end{tabular}
    \caption{Rephrasing quality evaluation average results. }
    \label{tab:para-qua}
\end{table}

As can be seen from the results, the rephrasing outputs have relatively high BERT-Score and human evaluation scores. This observation clearly demonstrates that using Llama-2-Chat-7B for rephrasing is suitable and does not interfere with contamination detection.

\paragraph{Performance Stability: Rephrasing}
We choose Llama-2-Chat-7B for rephrasing because it is a powerful model. However, the rephrasing model $M_p$ does not affect the final result as long as the model is capable enough. To validate our method provides stable results using different rephrasing models, we use another model, Mistral-v0.2-Instruct-7B \citep{jiang2023mistral}, for rephrasing. Other settings remain the same as in the previous experiments. The results are shown in Table \ref{tab:stab-para}.

\begin{table}[htbp]
    \centering
    \small
    \begin{tabular}{cccc}
    \toprule
        Data & \makecell[c]{Rephrase \\ Model} & \makecell[c]{Llama \\
        (Contaminated)} & \makecell[c]{Llama\\ (Original)} \\
        \midrule
        \multirow{2}{*}{\makecell[c]{Trained \\Data}} & Llama & \textbf{6e-8} & 0.12 \\
        ~ & Mistral & \textbf{2e-3} & 0.99\\
        \midrule
        \multirow{2}{*}{\makecell[c]{Untrained\\Data}} & Llama &  0.92 & 0.92\\
        ~ & Mistral & 0.23 & 0.99\\
        \bottomrule
    \end{tabular}
    \caption{p-value of different rephrase models. The significant results are in bold.}
    \label{tab:stab-para}
\end{table}

 Using either Llama or Mistral for rephrasing does not affect the outcomes, confirming that we can select any sufficiently powerful model for rephrasing. We use Llama-2-Chat-7B for rephrasing in our other experiments as mentioned earlier.


\paragraph{Performance Stability: Contamination Types}
As is discussed, there are two types of benchmark contamination. Our previous experiments primarily focus on the second type, as it involves shorter trained parts and is somewhat harder to detect. However, our method is also  capable of detecting the first type. The results are shown in Table \ref{tab:stab-type}.

\begin{table}[htbp]
    \centering
    \small
    \begin{tabular}{c|cc}
    \toprule
        Data & \makecell[c]{Llama  (Cont. I)} & \makecell[c]{Llama  (Cont. II)} \\
        \midrule
        \makecell[c]{Trained} & \textbf{4e-15} & \textbf{6e-8} \\
        \midrule
        \makecell[c]{Untrained}& 0.75 & 0.92\\
        \bottomrule
    \end{tabular}
    \caption{p-value of different contamination types. The significant results are in bold. Cont. represents contaminated.}
    \label{tab:stab-type}
\end{table}

As can be seen from the results, our method still works properly under the first contamination type. This result shows that our method is able to detect contamination with different types, which further proves its effectiveness.

\paragraph{Performance Stability: Randomness}
Paraphrasing unavoidably introduces randomness into contamination detection, so it is necessary to investigate the stability of our method under such conditions. We conduct this experiment using the same settings as above but randomly select five seeds to explicitly introduce randomness. The results are listed in Table  \ref{tab:stab:rand}.


With the random seed changing, the corresponding p-value changes, but the significance (or in-significance) remains unchanged. This observation clearly demonstrates the stability of our method under randomness.

\section{Prompt}

\label{sec:paraphrase prompt}
\begin{figure*}[ht]
\begin{tcolorbox}[title={Rephrase Prompt}, label=prompt:multi-stage-1]
\textbf{Instruction:} You are provided with a question. 
        Your task is to rephrase this question into another question with the same meaning.
            When rephrasing the question, you must ensure that you follow the following rules:  \\
            (1). You must ensure that you generate a rephrased question as your response. \\
            (2). You must ensure that the rephrased question bears the same meaning with the original question. Do not miss any information. \\
            (3). You must only generate a rephrased question. Any other information should not appear in your response. \\
            (4). Do not output any explanation. \\
            (5). Do not modify the numbers or quantities in the question. You should remain them unchanged
\\
\textbf{Example:}\\
\{In-Context Examples\}

\textbf{Input:} \\
\{input\}\\
\\
\textbf{Output:}\\

\end{tcolorbox}
\end{figure*}

\begin{figure*}[ht]
\begin{tcolorbox}[title={Confidence Estimation Prompt}, label=prompt:multi-stage-2]
\textbf{Instruction:} You are an expert in judging whether the answer is correct. You will be given a question and a corresponding answer. 
        Your job is to determine whether this answer is correct. 
        You should only respond with Yes or No.

\textbf{Example:}\\
\{In-Context Examples\}

\textbf{Input:} \\
\{input\}\\
\\
\textbf{Output:}\\

\end{tcolorbox}
\end{figure*}

\begin{figure*}[ht]
\begin{tcolorbox}[title={Confidence Estimation Example}, label=prompt:multi-stage-2-example]
\textbf{Instruction:} You are an expert in judging whether the answer is correct. You will be given a question and a corresponding answer. 
        Your job is to determine whether this answer is correct. 
        You should only respond with Yes or No.

\textbf{Example:}\\
\{In-Context Examples\}

\textbf{Input:} \\
The question is: At what concentration does
prolonged exposure to phosgene become
dangerous?
A. 100 ppm B. 25 ppm C. 1 ppm D. 10 ppm \\

The answer is B. \\

Is the answer correct according to the given question?
\\
\textbf{Output:}\\
Yes.

\textbf{Output Distribution:}
$P(Yes) = 0.92$, which means confidence $c=0.92$.
\end{tcolorbox}
\end{figure*}

\end{document}